\documentclass[letterpaper, 10 pt, journal, twoside]{IEEEtran}

\usepackage{cite}
\usepackage{amsmath,amssymb,amsfonts}
\usepackage{algorithmic}
\usepackage{graphicx}
\usepackage{textcomp}
\usepackage{xcolor}
\def\BibTeX{{\rm B\kern-.05em{\sc i\kern-.025em b}\kern-.08em
    T\kern-.1667em\lower.7ex\hbox{E}\kern-.125emX}}

\usepackage[symbol]{footmisc}
\usepackage{subcaption}
\usepackage{hyperref}
\usepackage{amssymb}
\usepackage{booktabs}
\usepackage{siunitx}
\usepackage[font=footnotesize]{caption}

\begin{document}

\title{Effects of Shared Control on Cognitive Load and Trust in Teleoperated Trajectory Tracking\\}

\author{Jiahe Pan$^{1}$, Jonathan Eden$^{1}$, Denny Oetomo$^{1}$, and Wafa Johal$^{1}$
\thanks{Manuscript received: December, 8, 2023; Revised March, 6, 2024; Accepted April, 14, 2024.}
\thanks{This paper was recommended for publication by Editor Angelika Peer upon evaluation of the Associate Editor and Reviewers' comments.
This work was partially supported by the Australian Research Council (Grant Nos. DE210100858 and DP240100938)}
\thanks{$^{1}$The authors are with the Faculty of Engineering and Information Technology, The University of Melbourne, Australia.
{\tt\footnotesize jmpan@student.unimelb.edu.au, \{eden.j, doetomo, wafa.johal\}@unimelb.edu.au}}%
\thanks{Digital Object Identifier (DOI): see top of this page.}
}

\markboth{IEEE Robotics and Automation Letters. Preprint Version. Accepted April, 2024}
{Pan \MakeLowercase{\textit{et al.}}: Effects of Shared Control on Cogn. Load and Trust}

\maketitle

\begin{abstract}
Teleoperation is increasingly recognized as a viable solution for deploying robots in hazardous environments. Controlling a robot to perform a complex or demanding task may overload operators resulting in poor performance. To design a robot controller to assist the human in executing such challenging tasks, a comprehensive understanding of the interplay between the robot's autonomous behavior and the operator's internal state is essential. In this paper, we investigate the relationships between robot autonomy and both the human user's cognitive load and trust levels, and the potential existence of three-way interactions in the robot-assisted execution of the task. Our user study (N=24) results indicate that while the autonomy level influences the teleoperator's perceived cognitive load and trust, there is no clear interaction between these factors. Instead, these elements appear to operate independently, thus highlighting the need to consider both cognitive load and trust as distinct but interrelated factors in varying the robot autonomy level in shared-control settings. This insight is crucial for the development of more effective and adaptable assistive robotic systems.
\end{abstract}

\begin{IEEEkeywords}
Human-Robot Collaboration, Acceptability and Trust, Telerobotics and Teleoperation
\end{IEEEkeywords}

\section{Introduction} \label{sec:introduction}
\IEEEPARstart{F}{rom} teleoperating a robot to inspect a disaster site to using a rehabilitation robot after stroke, human-robot interaction (HRI) aims to combine the respective strengths of humans and robots to achieve desired task outcomes. However, how to best configure robot autonomy during HRI such that it improves performance without being negatively perceived by its user remains an open problem. When a user and a robot work collaboratively to complete a task, problems such as unsafe operation or reduced user performance can arise if the user distrusts the robot and its behaviors \cite{kok2020trust}, or if they have a sufficiently large cognitive load such that they cannot reliably command the robot \cite{baltrusch2022human}.

The main research question of this study is: \textit{How does the user's cognitive load and trust vary with changes in the robot autonomy level during HRI?} To answer this question, we conducted a teleoperation study using a shared-control scheme that allowed for varying the robot autonomy level \footnote[2]{A demo video is available at \href{https://youtu.be/yizsBG1QJog}{https://youtu.be/yizsBG1QJog}, and the code and data are found at \href{https://github.com/mpan31415/AutonomyCLTrust}{https://github.com/mpan31415/AutonomyCLTrust}.}. This used a trajectory tracking task to minimize the effects of confounding variables while enabling a fundamental understanding of robot autonomy's effect on human cognitive load and trust. Through metrics that captured trust and cognitive load, via both objective methods and questionnaires, we looked to identify the relationship between the autonomy level and these factors. Specifically, we formulate three main hypotheses:
\begin{itemize}
    \item \textbf{H1} - The participants' cognitive load will decrease with increasing robot autonomy
    \item \textbf{H2} - The participants' trust in the robot will increase with increasing robot autonomy
    \item \textbf{H3} - There will be an autonomy dependent relationship between trust and cognitive load
\end{itemize}

\section{Related Works} \label{sec:related_works}

Teleoperation refers to robot operation using human intelligence via a human-machine interface \cite{cui2003review}, and has shown robustness and real-world applicability in assisting humans to perform complex tasks in hazardous and uncertain environments \cite{alvarez2001reference}. As both robot and human intelligence are often required to successfully complete these tasks, teleoperation often makes use of shared-control \cite{kent2017comparison}, where command inputs from the human and the robot's autonomous controller are integrated and arbitrated to determine the robot's resulting actions. Here, existing shared-control methods have mostly focused on the task performance and/or the physical characteristics of the interaction such as the human and/or robot effort. These interaction metrics are also often task-specific, such as minimizing collisions with obstacles while navigating \cite{deng2019bayesian} or maximizing the percentage of successful object grasps \cite{zhuang2019shared}. 

It has however been shown that robot autonomy and behavior can also impact the human user's internal state, such as their cognitive load \cite{baltrusch2022human, martinetti2021redefining} and trust \cite{kok2020trust, langer2019trust}. Cognitive load is as an indicator of task complexity based on the number of conceptual elements that need to be held in the user's mind at any one time to solve a specific task \cite{plass2010cognitive}. Here, subjective assessment schemes (e.g., Subjective Workload Assessment Technique (SWAT) \cite{SWAT}, Rating Scale of Mental Effort (RSME) \cite{RSME}, Task Load Index (NASA-TLX) \cite{NASA-TLX}) have been shown to reliably capture the perception of workload \cite{marchand2021measuring} but can only be conducted at low frequencies. In contrast, the dual-task paradigm of performing a secondary task (e.g., reproducing melodies \cite{park2015rhythm, sun2016probing}, performing simple math \cite{lee2015influence, tang2015motor}, responding to stimuli \cite{scerbo2017differences, wirzberger2018schema}) concurrently with a primary task and physiological measures (e.g., pupil dilation \cite{eckstein2017beyond, white2017usability}, cardiac activity \cite{collet2009autonomic, mehler2009impact}, brain activity \cite{miller2011novel, klimesch1999eeg}) allow the objective measure of cognitive load at higher frequencies \cite{marchand2021measuring} but suffer from the impact of confounding factors such as learning effects \cite{esmaeili2021current} and expertise in multitasking \cite{strobach2015better}.

A human's trust in another agent is defined as a multidimensional latent variable that mediates the relationship between events in the past and the former agent's subsequent choice of relying on the latter in an uncertain environment \cite{kok2020trust}. Trust is therefore an internal measure experienced by humans, making it difficult to capture objectively \cite{chita2021can}. Existing methods that have been explored include physiological information \cite{kohn2021measurement} and brain imaging \cite{wang2018eeg} which do not necessarily capture trust independent of confounding factors. As a result, subjective questionnaires such as the Multi-Dimensional Measure of Trust (MDMT) \cite{mdmt1} which captures trust across 2 categories - \{\textit{Capacity Trust, Moral Trust}\} are most widely employed in HRI studies \cite{ahmad2022no, ullman2021challenges}.

Existing works have only studied the effect of autonomy on either the user's cognitive load \cite{lin2020shared, young2019formalized} or trust \cite{luo2021workload} separately. However, there is evidence that cognitive load and trust may not be independent measures \cite{chen2016trust, ahmad2019trust}. Therefore, it is important to understand whether cognitive load and trust vary based on an underlying dependency between them, or if they are in fact independently influenced by autonomy. 

To the best of our knowledge, there is currently no reported study on the three-way relationship between robot autonomy, cognitive load, and trust. Having a fundamental understanding of such a relationship is crucial, as with it roboticists can better design adaptive robots with varying autonomy levels that adjust their behavior in response to their user. Such adaptive schemes are gaining interest in areas like autonomous driving \cite{luo2021workload} and personal service robots \cite{preusse2020m}.

\begin{figure}
\centering
\begin{subfigure}{0.24\textwidth}
    \includegraphics[width=\textwidth]{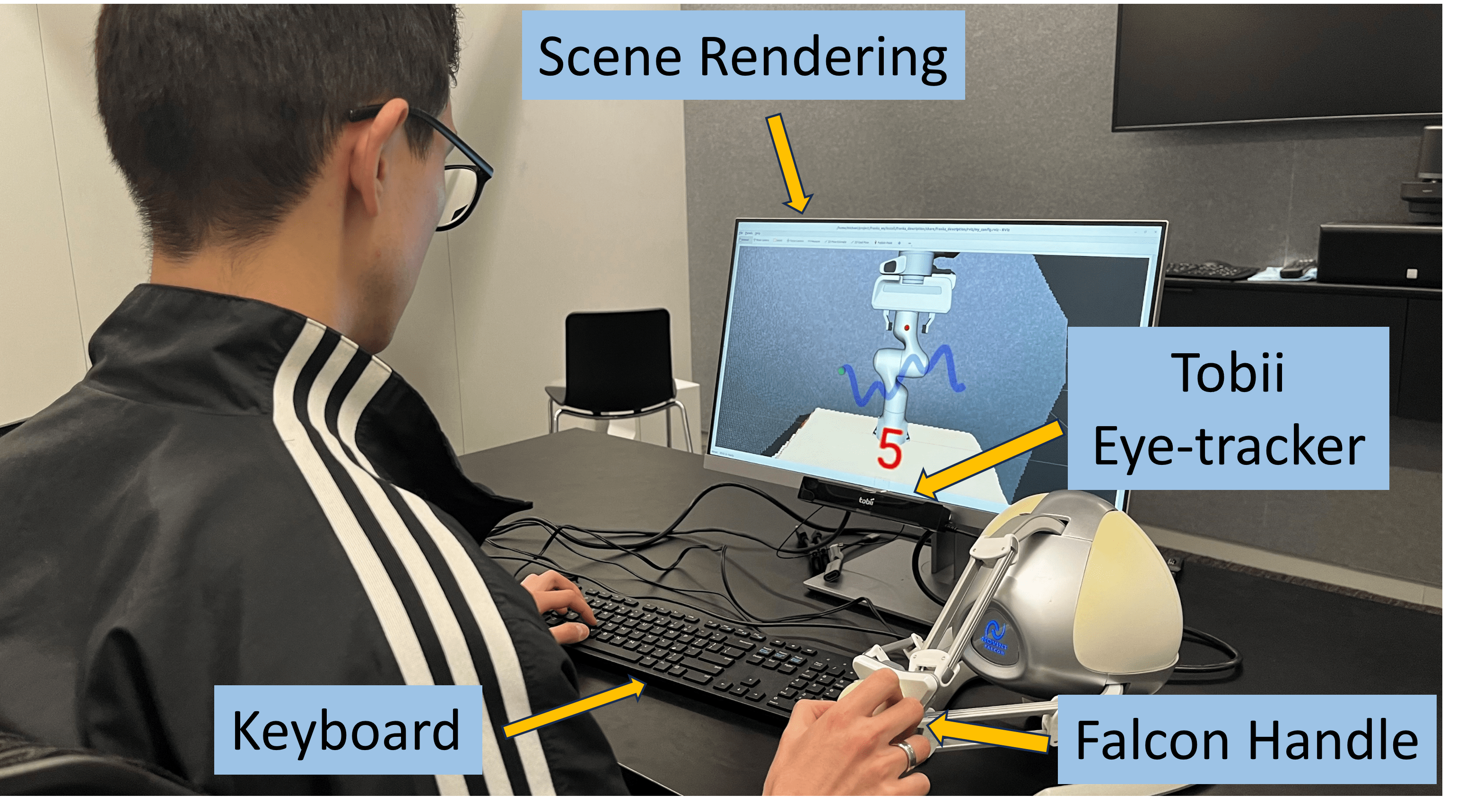}
\end{subfigure}
\begin{subfigure}{0.24\textwidth}
    \includegraphics[width=\textwidth]{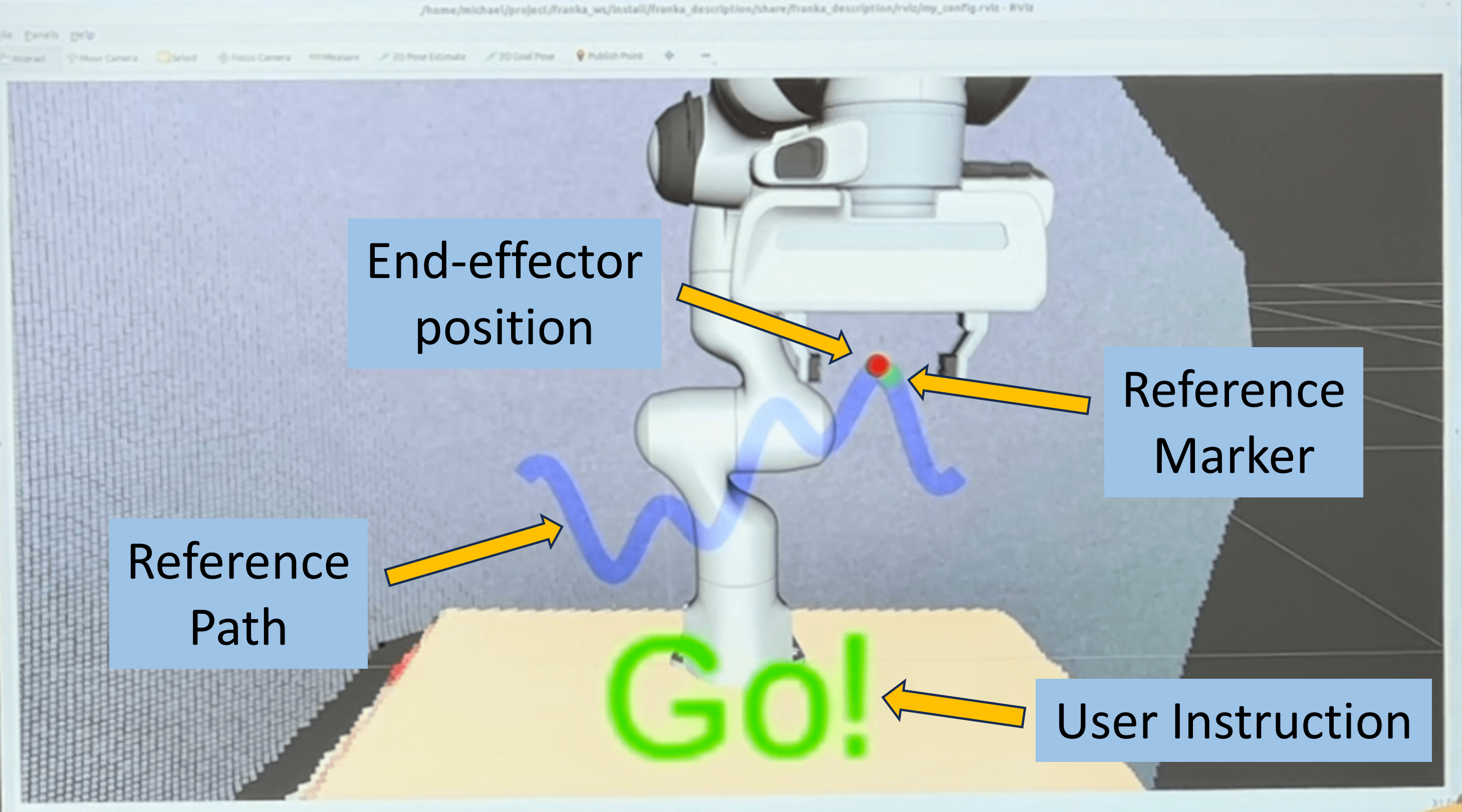}
\end{subfigure}
\caption{Experiment setup. \textbf{(Left)} Task setup. The primary trajectory-tracking task was performed via a Novint Falcon controller and visual feedback on the screen. The secondary task involved using the ``space key'' to tap a given rhythm. \textbf{(Right)} Primary task visualisation. The trajectory to be followed was displayed in blue. At each time instant, the red dot showed the position of the center of the end-effector and the green dot showed the target.}
\label{fig:entire_setup}
\end{figure}

\section{Method} \label{sec:methodology}

Our study evaluated the cognitive load and trust of participants during their teleoperation of a (Franka Emika Research 3) robotic arm. The task entailed tracking a 3-dimensional trajectory with the robot's end-effector. Trajectory tracking was selected as the primary task as it is a component within a variety of real-world applications, such as robotic welding \cite{ogbemhe2015towards, lei2020review}, cutting \cite{chen2013robot, yuwen2022path} and painting \cite{chen2017paint}. Furthermore, it enabled the use of clearly defined robot autonomy levels within the blending control scheme (see Section \ref{subsec:primary_task}) and has a clear metric on task performance through the well-established root mean square error (RMSE) metric (see Section \ref{subsec:measures}). In addition to this primary task, we employed a dual-task methodology to measure cognitive load, with a secondary rhythmic tapping task.

\subsection{User Interface} \label{subsec:user_interface}
The user interface was designed to replicate a real-world teleoperation scenario, where the participant operated from a distinct environment with no direct perception of the true physical robot, which was situated in an adjacent room. Instead, we presented a virtual rendering of the robot's real-time state and a pre-captured Point Cloud of its surrounding environment on a computer screen (Figure \ref{fig:entire_setup} Left) using Rviz \cite{kam2015rviz}. Teleoperation was performed via the handle of a Novint Falcon haptic controller, which was connected to the robot in the adjacent room via a local Ethernet connection. No haptic feedback was provided to participants in this study. Participants performed the secondary tapping task by pressing ``space" on a computer keyboard (Figure \ref{fig:entire_setup} Left).

\subsection{Primary Task} \label{subsec:primary_task}
The primary task required the participant to teleoperate the robot to track a trajectory (see Figure \ref{fig:entire_setup} Right) of  \(T = 10\)\,s duration using the robot's end-effector. Participants were asked to ``as accurately as possible'' follow the 3-dimensional time-parameterized trajectory with \(t \in [0, T]\)
\begin{equation}
\begin{aligned}
    x(t) &= x_0 \\
    y(t) &= y_0 + d_y (t/T) \\
    z(t) &= z_0 + d_z f(t),
\end{aligned}
\label{eqn:time_traj}
\end{equation}
where 
\begin{equation}
    f(t) = sin(a_1(t+\theta)) + sin(a_2(t+\theta)) + sin(a_3(t+\theta))
    \label{eqn:multi-sine}
\end{equation}
is a multi-sine curve with parameters \(a_1, a_2, a_3 \in \mathbb{N}\) and \(\theta \in \mathbb{R}\). Here, \((x_0, y_0, z_0)\) represents a constant offset from the robot's base, and \(d_y = 0.3\)\,m and \(d_z = 0.15\)\,m are the width and height parameters of the trajectory. Using \eqref{eqn:time_traj} and parameters \((a_1, a_2, a_3, \theta)\), we generated a set of trajectories \(S = \{ (2,3,4,\frac{4\pi}{3}), (1,3,4,\pi), (2,2,5,\pi),(2,3,5,\frac{8\pi}{5}), (2,4,5,\pi)\}\). These different trajectories were used to make it unlikely for memorization during multiple rounds of repetition. The parameters were chosen with similar amplitudes and frequencies to ensure similar tracking complexity. 

The robot was commanded through shared-control between the human and its own autonomous controller. This adopted a blending control (or convex combination) formulation \cite{dragan2013policy,marcano2020review}, where the input position was computed as
\begin{equation}
\begin{aligned}
    \mathbf{u} = \gamma \mathbf{u_r} + (1-\gamma) \mathbf{u_h}, \quad
    \gamma &\in [0, 1] \subset \mathbb{R}.
\end{aligned}
\label{eqn:std_convex}
\end{equation}
Here \(\mathbf{u_r} \in \mathbb{R}^3\) denotes the reference robot end-effector position input which corresponded to the trajectory value at that time, and \(\mathbf{u_h} \in \mathbb{R}^3\) denotes the reference end-effector position from the human's input - determined by a constant 3 times scaling of the Falcon handle position, which was read at 500\,Hz. The scalar \(\gamma\) represents the level of \textit{robot autonomy}, where \(\gamma = 1\) corresponds to complete robot control (\textit{full autonomy}) and \(\gamma = 0\) corresponds to complete human control (\textit{no autonomy}). By altering \(\gamma\) the relative autonomy between the human and robot was varied. The resulting input \(\mathbf{u} \in \mathbb{R}^3\) was passed in ROS to the Kinematics and Dynamics Library (Orocos) inverse-kinematics solver to compute joint positions, which were then tracked by the robot's joint controllers running at 500\,Hz.

\subsection{Secondary Task} \label{subsec:secondary_task}
\begin{figure}[t!]
    \centering
    \includegraphics[width=0.48\textwidth]{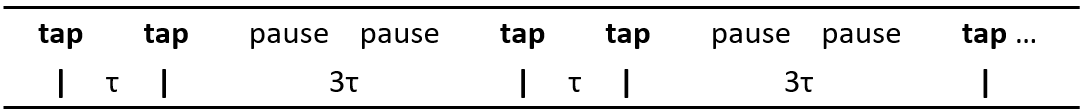}
    \caption{Rhythm Method Visualization (in four-four time). Participants followed the rhythm by tapping the ``space" bar. The repeating pattern consisted of a short inter-tap interval followed by a long one (adapted from \cite{park2015rhythm}).}
    \label{fig:rhythm_method}
\end{figure}

The secondary task consisted of the ``Rhythm Method'' \cite{park2015rhythm}. This involved the participant tapping a pre-recorded rhythm from memory while concurrently performing the primary task. Our implementation followed that of \cite{park2015rhythm}, except that we generated multiple rhythms that used different tempos with the same pattern. Similar to our trajectory design, this was done to minimize the chance of participants learning the rhythm across multiple rounds. As shown in Figure \ref{fig:rhythm_method}, the pattern consisted of a short inter-tap interval of duration \(\tau\), followed by a long inter-tap interval of duration \(3 \tau\). Hence, the speed of the rhythms were inversely related to and directly determined by the duration of the short inter-tap interval, \(\tau\). Using this parametrization, we generated 5 rhythms of different speeds with \(\tau \in [400, 600]\)\,ms. For consistency with a dual-task paradigm, we explicitly instructed the participants that the primary task always \textit{takes priority} over the secondary task. 

Using this secondary tapping task alongside the primary teleoperation task, we aimed to capture the participant's cognitive load and gain insights into how they can effectively manage their cognitive resources when faced with concurrent demands. This mirrored situations in which individuals need to multitask or divide their attention in real-world scenarios.

\subsection{Measures} \label{subsec:measures}
We used objective and subjective measures to investigate the relationship between the robot's \textit{autonomy level} in the teleoperation task and the participant's cognitive load and trust.

\subsubsection{Demographics Information} \label{subsubsec:demo_info}
For each participant, we collected basic demographic information via an initial questionnaire at the start of the session. Additionally, due to their potential relevance during teleoperation, we included the following three items which participants rated on a 10-point Likert scale from 1 (\textit{strongly disagree}) to 10 (\textit{strongly agree}): i) I trust new technology in general; ii) I play computer games regularly; and iii) I am proficient in a musical instrument.

\subsubsection{Performance and Physiological Measures} \label{subsubsec:in_task_measures}
During each trial, we collected data pertaining to the participant's performance on the primary and secondary tasks, along with eye-tracking data. This was used to evaluate cognitive load. 

\textbf{Trajectory Tracking Error} - The robot position information during each trial was recorded at \(20\)\,Hz. The trajectory tracking error was calculated as the RMSE between the reference and recorded trajectories. Feedback from initial pilot studies showed that participants found it difficult to perceive and track depth components using the RViz display. Therefore, the RMSE calculation of the trajectory tracking error did not include the error in tracking the constant depth.

\textbf{Tapping Error} - Following \cite{park2015rhythm}, we recorded the time stamp of each tap executed by the participant. We then calculated the duration of each of the inter-tap intervals separately for both interval types (short and long inter-tap intervals). Finally, the tapping error was calculated by subtracting the average baseline error from the average recorded error for each trial and then normalizing with respect to the interval lengths.

\textbf{Pupil Diameter} - Larger pupil dilation has been associated with more intense cognitive processing \cite{eckstein2017beyond, white2017usability}. Therefore, we recorded pupil diameter data during each trial at 40 Hz using the Tobii Pro X2 screen-based eye-tracker. To derive the cognitive load estimate from this data, we applied the method outlined in \cite{duchowski2020low}, which involves performing a wavelet decomposition of the data sequence and computing a ratio of low/high frequencies of pupil oscillation.

\subsubsection{Self-Reported Measures} \label{subsubsec:questionnaire_measures}
After each round of five trials, the following measures were collected via questionnaires:

\textbf{Perceived Autonomy} - The ``perceived autonomy questionnaire" \cite{harbers2017perceived} has been widely used in HRI \cite{roesler2022influence, balatti2020method} to capture user perception of autonomy. Adapting the questionnaire, we included the item: \textit{``How autonomous did you feel the robot was?"}, which participants reported on a 10-point discrete scale. This self-reported measure acted as a manipulation check and assessed if participants perceived the change of the robot's autonomy level between rounds. 

\textbf{Cognitive Load} - We administered the NASA-TLX questionnaire \cite{NASA-TLX} with all six original sub-scales to provide a comprehensive assessment of cognitive workload.

\textbf{Trust} - The MDMT questionnaire \cite{mdmt1} is composed of 8 sub-scales in each of the \textit{Capacity Trust} and \textit{Moral Trust} categories. Since our study design focused on the human's trust relating to the robot's behavior in the trajectory tracking task, we chose to only include the 8 sub-scales of \textit{Capacity Trust}, and excluded all items in the \textit{Moral Trust} category.

We complemented the MDMT with a single-scale trust question: \textit{``How much do you trust the robot?"}, which participants also reported using a 10-point discrete scale. This was included to see if a single trust measure could be reliably used as a high-frequency self-reported measure instrument.

\subsection{Study Design} \label{subsec:study_design}

\subsubsection{Conditions} \label{subsubsec:conditions}
To test our hypotheses, we used a counterbalanced, within-subject design to examine the effects of robot autonomy while controlling for order effects. The within-subject condition was the robot's \textit{autonomy} level, which we designed to be either \textit{high} (\(\gamma=\{0.6, 0.8\}\)) or \textit{low} (\(\gamma=\{0.0, 0.2\}\)). Two blocks of trials were conducted using the corresponding two values of \(\gamma\) for each \textit{autonomy} condition. The balanced experimental design meant that half of the participants were exposed to these two autonomy levels in \textit{increasing} (\(\gamma \in \) \textit{low} then \textit{high}) order, and the other half in \textit{decreasing}  (\(\gamma \in \) \textit{high} then \textit{low} ) order. 

Using the experimental conditions, we performed an a priori power analysis to determine our sample size using G*Power \cite{faul:2007}. Considering a repeated-measure ANOVA with 0.8 power, \(\alpha\) = 0.05 and a medium effect size of \(f\) = 0.25, the calculation resulted in a sample size of \(N = 24\) (split evenly into the two groups).

\subsubsection{Participants} \label{subsubsec:participants}
Ethical approval for the study was granted by The University of Melbourne's Human Ethics Committee under project ID 27750. The 24 participants were aged from 18 to 32 years (\(M = 23.5, SD = 3.88)\), 18 identified as female and 6 as male. All participants were asked to perform the primary task using their preferred dominant hand (22 right-handed). Participants received a \$20 gift voucher as compensation at the end of the approximately 60 minute study. 

\begin{figure}[h]
    \centering
    \includegraphics[width=0.5\textwidth]{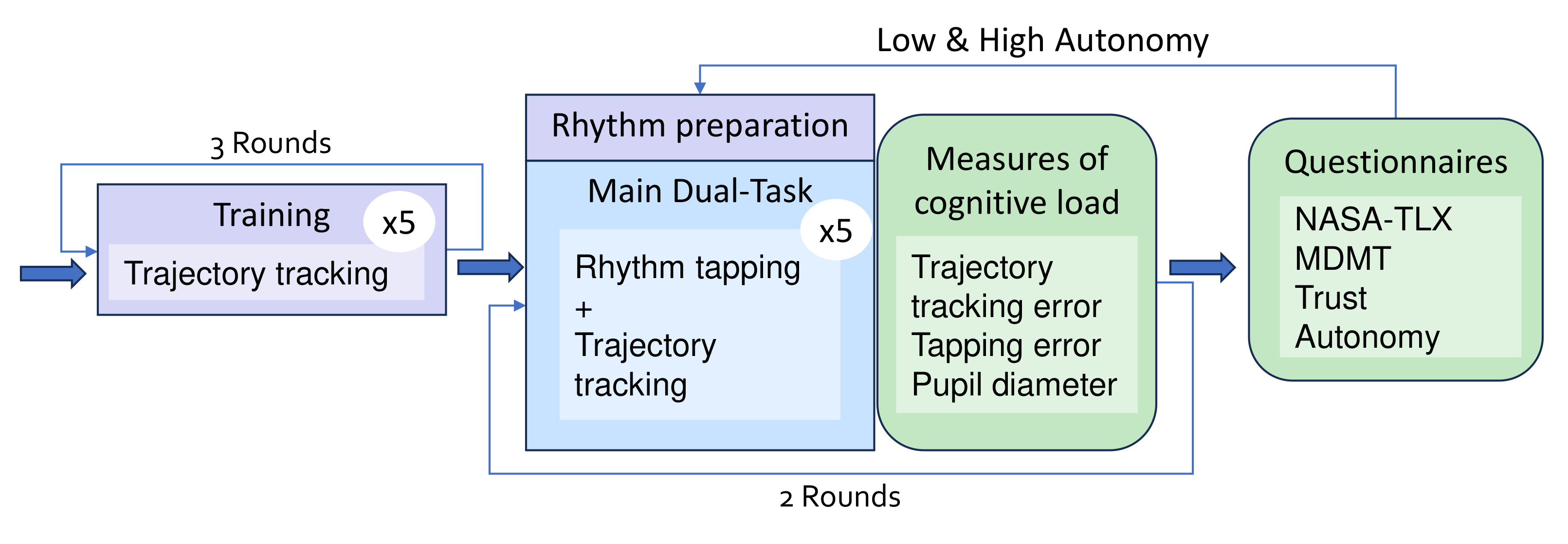}
    \caption{Experimental procedure including the training phase and main phase for a given autonomy level $\gamma$.}
    \label{fig:flowchart}
\end{figure}

\subsection{Experimental Procedure} \label{subsec:exp_procedure}
Each session had two phases (Figure \ref{fig:flowchart}): \textit{training} and \textit{main}. Within each phase, participants completed three rounds in training and two in main - where each round contained five trials - \{\(t_1, ..., t_5\)\}. During each trial, the task performance and pupil diameter data was continuously recorded (see Section \ref{subsubsec:in_task_measures}), while participants self-reported their cognitive load and trust after each round (see Section \ref{subsubsec:questionnaire_measures}). 

\subsubsection{Training Phase} \label{subsubsec:training_phase}
Participants were initially shown the shared-control setup and the visualization (see Figure \ref{fig:entire_setup}), followed by two practice trials for them to familiarize with the system and the trajectory tracking task. They were then trained to perform the tracking task in a single \textit{baseline} round. This round was used as a manipulation check to ensure that there was no effect of the secondary task on primary task performance and therefore required participants to perform the primary tracking task with \textit{no autonomy} and without the secondary tapping task. After, participants performed two rounds of both the tracking and tapping tasks, at a \textit{medium autonomy} level and a \textit{no autonomy} level, to familiarize with the dual-task scheme and for all participants to have a common baseline for our comparisons. Each round also included training for the tapping task, following the procedure of \cite{park2015rhythm}. In the training phase, participants were explicitly informed about the autonomy conditions of each round, which they could use as a reference for perceiving the robot's autonomy levels.

\subsubsection{Main Phase} \label{subsubsec:main_phase}
The two rounds of the main phase required participants to perform both teleoperation and tapping. Although they were informed that the robot's autonomy level might change from round to round, participants were not told the robot autonomy level, nor any information on its ability to perform the task. In each round, a different rhythm was used according to a pre-generated random-ordering. Participants were trained for tapping the rhythm, including a recorded section used as their baseline tapping performance. They then performed the five trials of the round, after which they completed the questionnaire to report their own subjective assessments of cognitive load and trust during the round. 

In each trial, there was an initial 5\,s preparation window shown as a 5\,s countdown on the screen (Figure \ref{fig:entire_setup}), during which participants heard a short recording of the rhythm. One of the five pre-generated trajectories was also displayed in a random-ordering, which included a blue line indicating the \textit{reference path} to track, and a green \textit{reference marker} which followed the path from start to finish over the 10\,s trajectory duration (see Section \ref{subsec:primary_task}). In addition, the robot's end-effector position was shown as a red dot located between its parallel grippers. After the preparatory 5\,s, the countdown text changed to the word \textit{``Go!"}, indicating that the trial has started. Participants were instructed to only perform both tasks during the 10\,s trajectory duration, after which the on-screen text changed to \textit{``Stop!"}, indicating that the trial had finished. Each session took approximately 60 minutes, including the 10-minute introduction block and a 10-minute block for each autonomy condition across both the training and main phases.

\section{Results} \label{sec:results}
Preliminary analysis showed no order effects on any measure. Therefore all subsequent analysis combined the data from the \textit{increasing} and \textit{decreasing} autonomy groups. After all measures were verified to be normally distributed using the Shapiro-Wilk test, we performed repeated-measures ANOVA with \textit{autonomy} as the within-subject variable. Results were considered significant at the threshold \(\alpha < .05\). The ANOVA results are summarized in Table \ref{table:anova}.

\begin{center}
\begin{table}[h!]
\centering
\begin{tabular}{lSSSSl}
\toprule
\textbf{Measure}                & {\textbf{(DFn, DFd)}} & {\textbf{F}}      & {\textbf{p}}     & {\textbf{$\eta^2$}} \\
\midrule
Perceived Autonomy     & {(1, 22)}    & 62.104   & <.001   & .262       \\
NASA-TLX               & {(1, 22)}    & 46.615   & <.001   & .262       \\
Tapping Error (short)  & {(1, 22)}    & 1.137    & .298    & .028       \\
Tapping Error (long)   & {(1, 22)}    & .873     & .36     & .008       \\
Pupil Diameter Index   & {(1, 22)}    & .088     & .769    & .0001      \\
MDMT                   & {(1, 22)}    & 41.418   & <.001   & .234       \\
Traj Tracking Error    & {(1, 22)}    & 129.005  & <.001   & .748       \\
\bottomrule
\end{tabular}
\caption{Summary of ANOVA results for all measures against \textit{autonomy}. DFn and DFd are the degrees of freedom in the numerator and denominator respectively, p specifies the p-value, $\eta^2$ is the generalized effect size.}
\label{table:anova}
\end{table}
\end{center}
\vspace{-0.5cm}
The demographics items listed in Section \ref{subsubsec:demo_info} were included as potential covariates. However, preliminary analysis showed no clear effect of these items on any of our dependent variables. Therefore, they were not included in the main analysis. Furthermore, potential learning effects across the five trials within each round were checked using Linear Mixed Models. While no learning effect was observed for any measure under both high and low \textit{autonomy} conditions, only the second round of each autonomy condition ($\gamma=0.8$ for \textit{high} autonomy and $\gamma=0.2$ for \textit{low} autonomy) was used in the analysis to eliminate any adaptation to a new autonomy level. For the same reason, the first trial data of each round was also removed from the analysis.

\begin{figure*}[t!]
    \centering
    
    \hspace*{\fill}
    \begin{subfigure}[t]{0.235\textwidth}
        \centering
        \includegraphics[width=\textwidth]{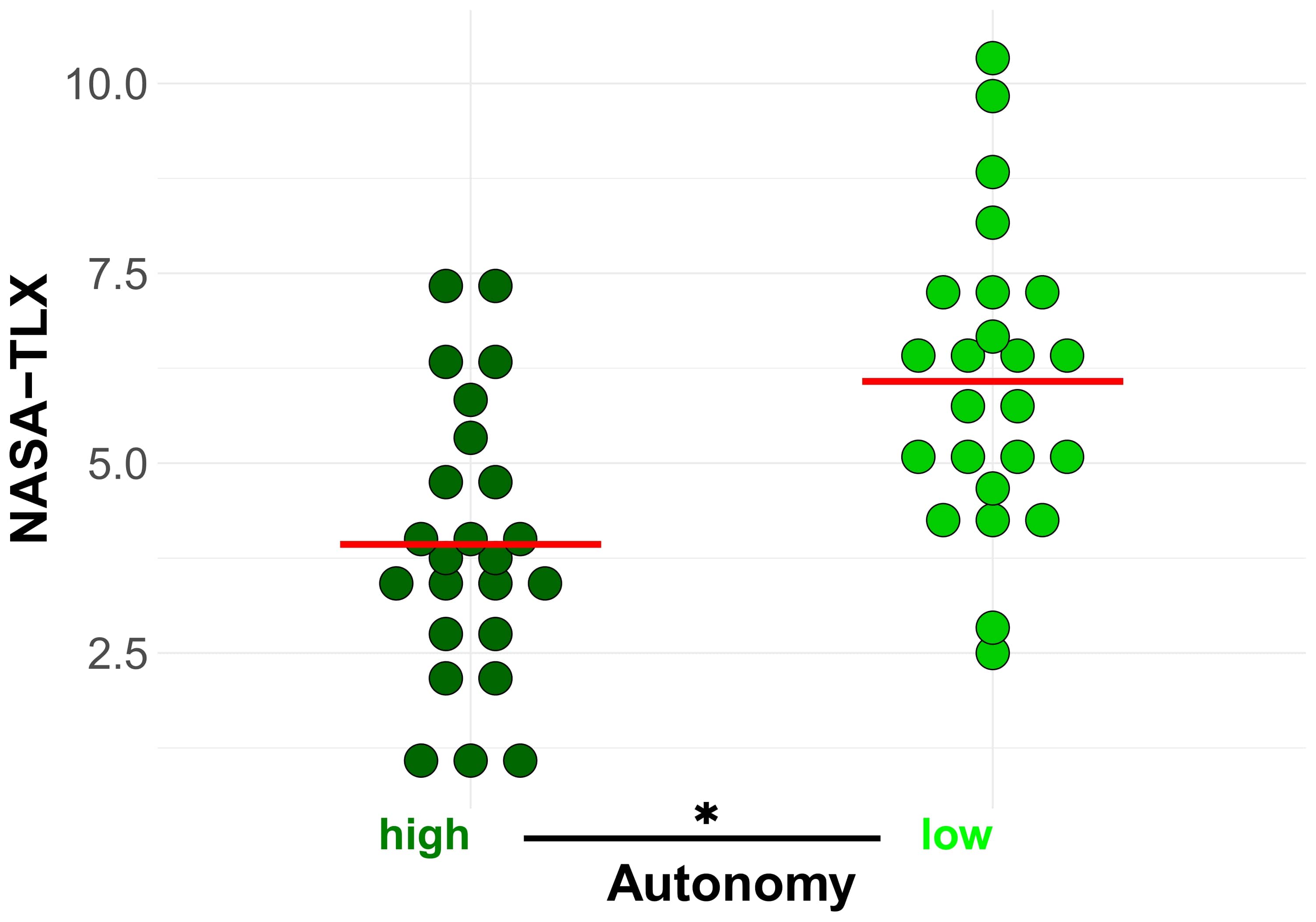}
        \caption{NASA-TLX}
        \label{fig:tlx}
    \end{subfigure}
    \hspace*{\fill}
    \begin{subfigure}[t]{0.235\textwidth}
        \centering
        \includegraphics[width=\textwidth]{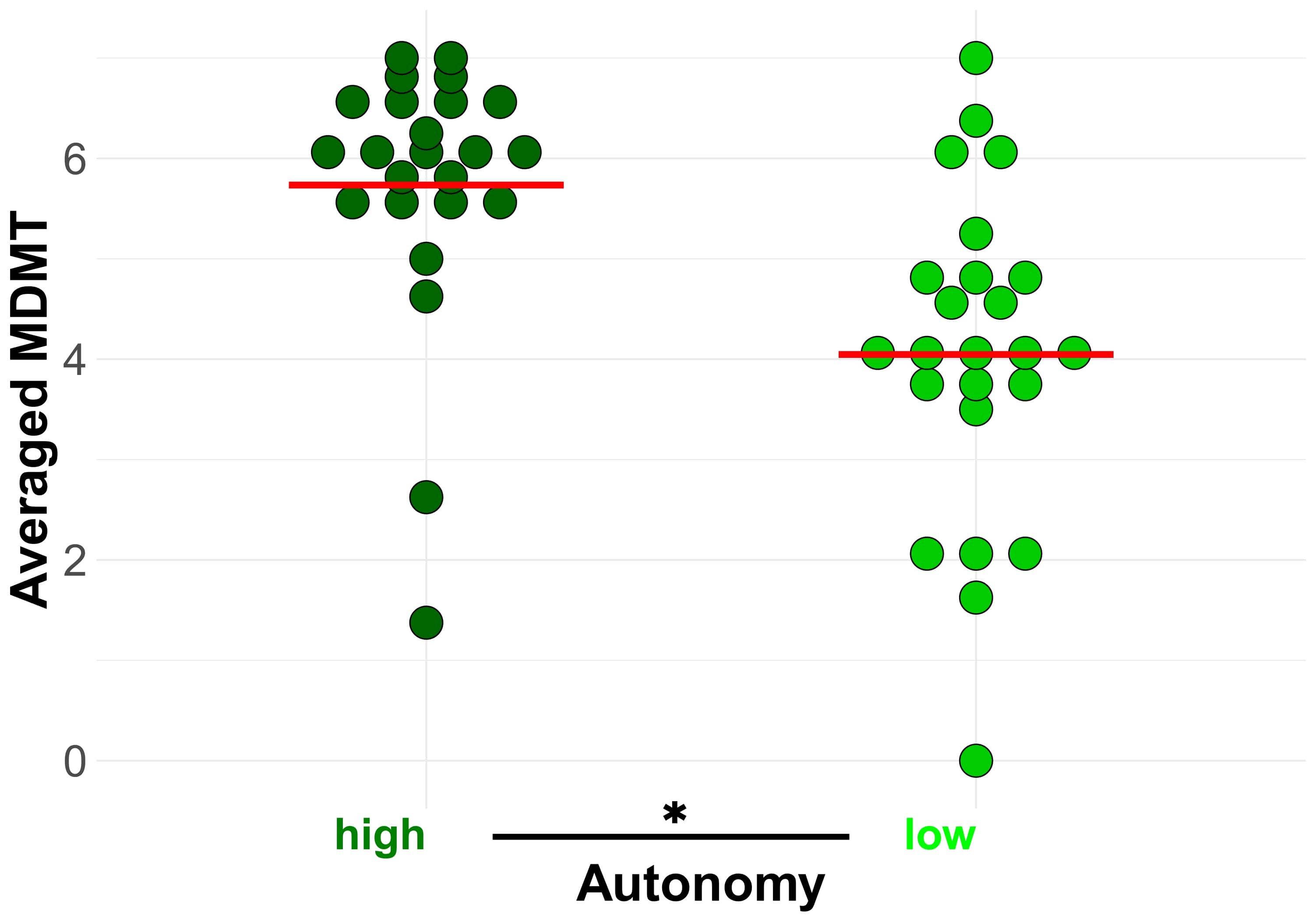}
        \caption{Averaged MDMT}
        \label{fig:mdmt}
    \end{subfigure}
    \hspace*{\fill}
    \begin{subfigure}[t]{0.235\textwidth}
        \centering
        \includegraphics[width=\textwidth]{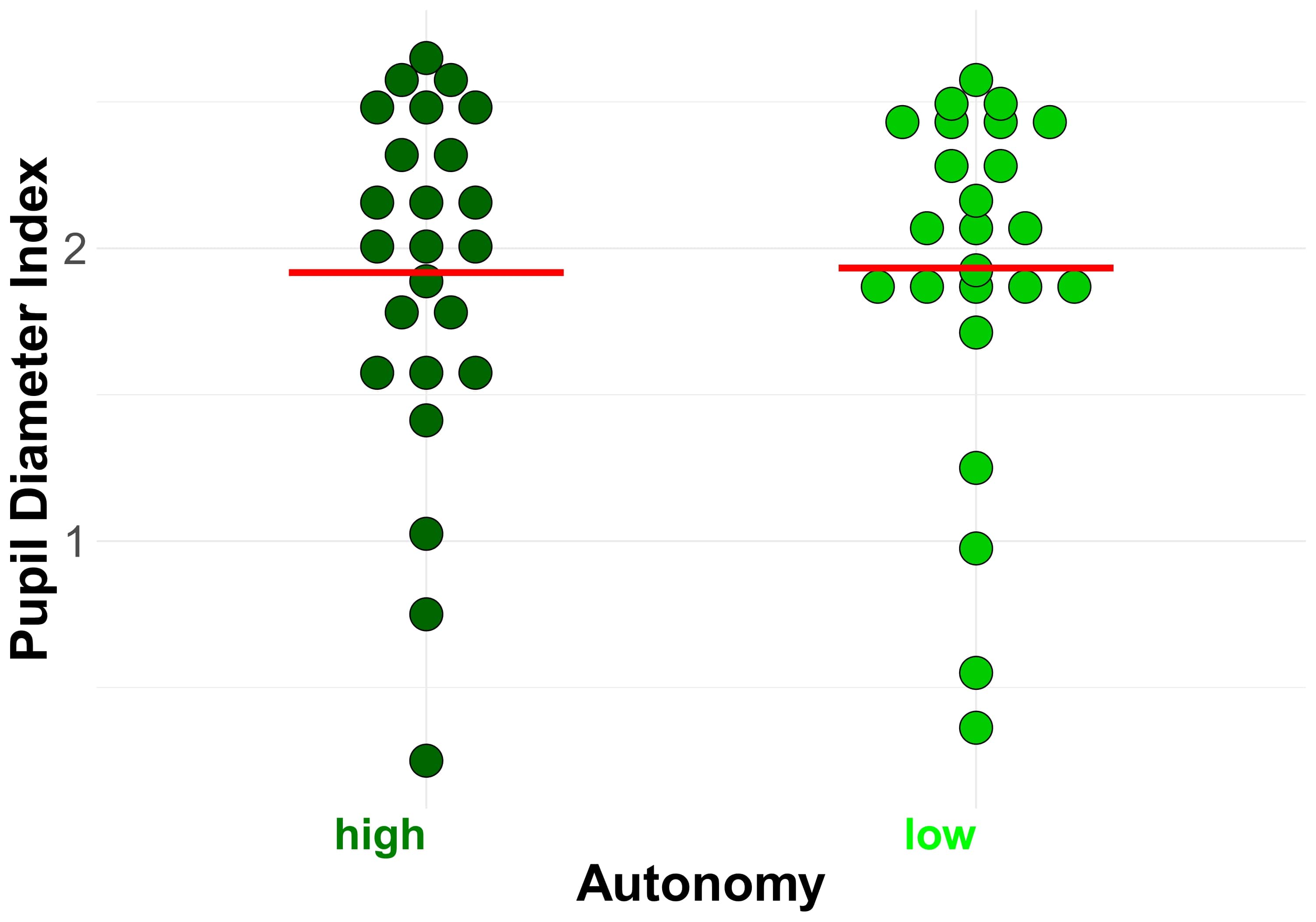}
        \caption{Pupil Diameter Index}
        \label{fig:pupil_diameter}
    \end{subfigure}    
    \hspace*{\fill}
    \begin{subfigure}[t]{0.235\textwidth}
        \centering
        \includegraphics[width=\textwidth]{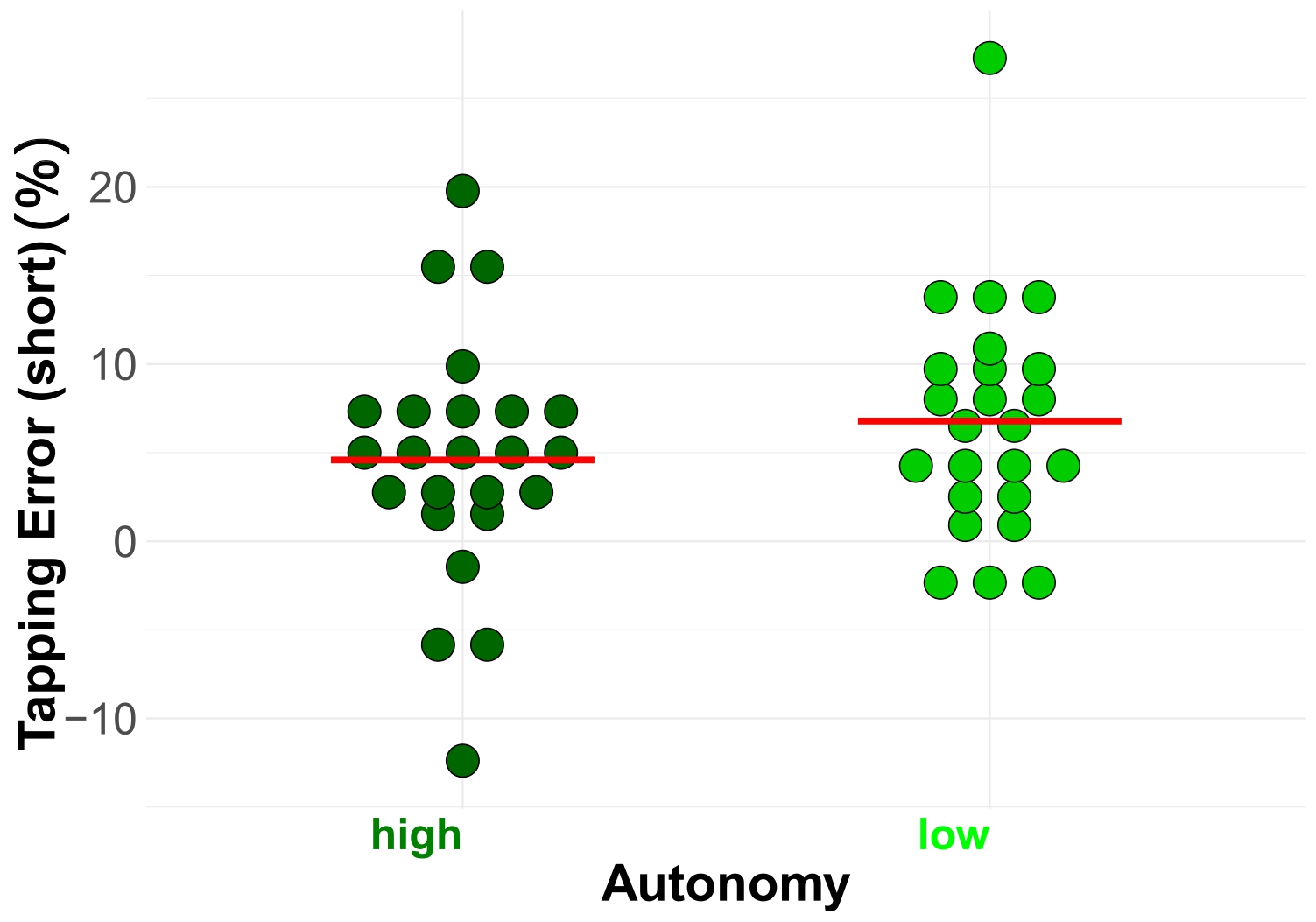}
        \caption{Tapping Error (short)}
        \label{fig:tapping_err_short}
    \end{subfigure}
    \hspace*{\fill}

    \medskip
    
    \hspace*{\fill}
    \begin{subfigure}[t]{0.235\textwidth}
        \centering
        \includegraphics[width=\textwidth]{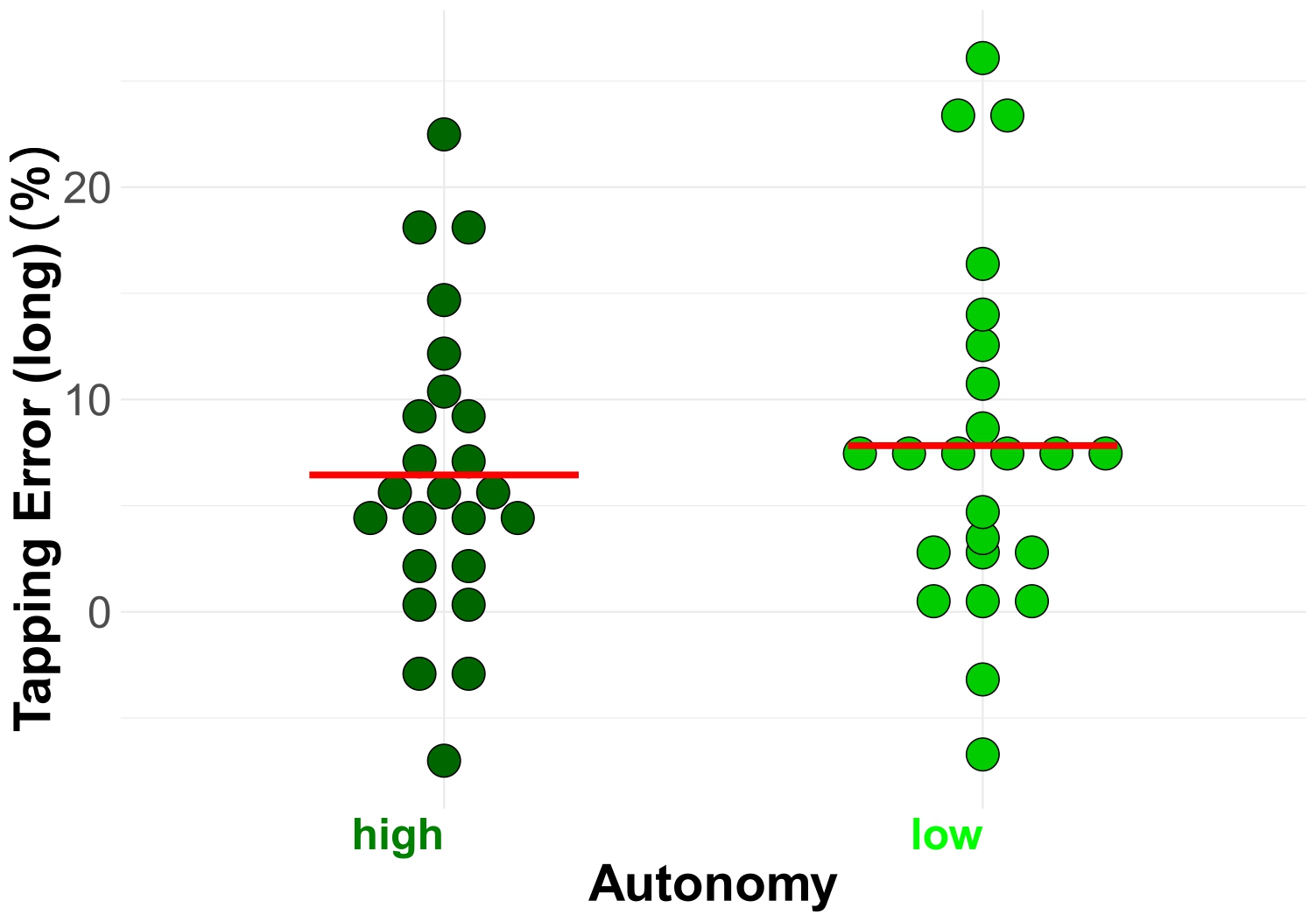}
        \caption{Tapping Error (long)}
        \label{fig:tapping_err_long}
    \end{subfigure}
    \hspace*{\fill}
    \begin{subfigure}[t]{0.235\textwidth}
        \centering
        \includegraphics[width=\textwidth]{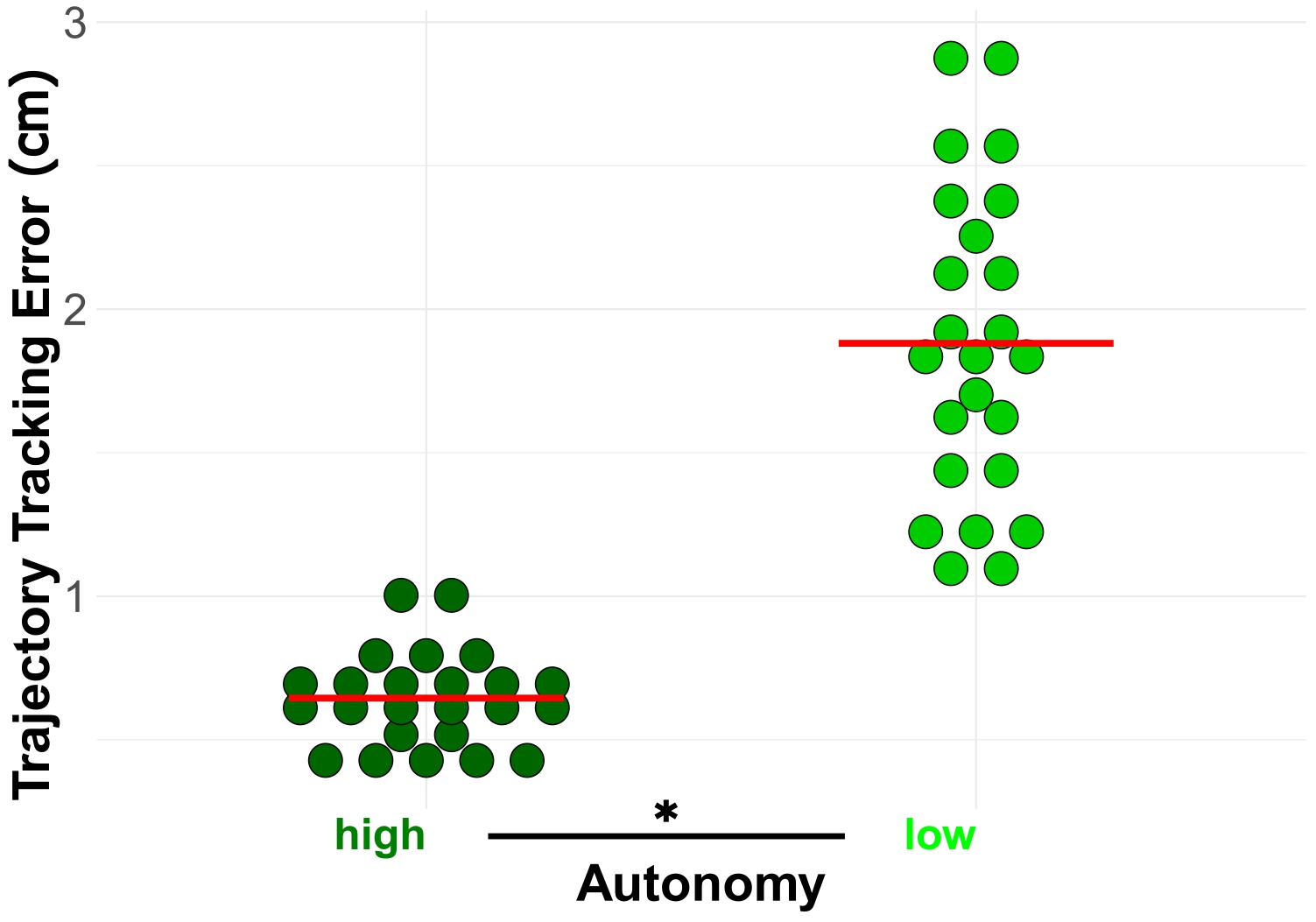}
        \caption{Trajectory Tracking Error}
        \label{fig:traj_err}
    \end{subfigure}
    \hspace*{\fill}
    \begin{subfigure}[t]{0.235\textwidth}
        \centering
        \includegraphics[width=\textwidth]{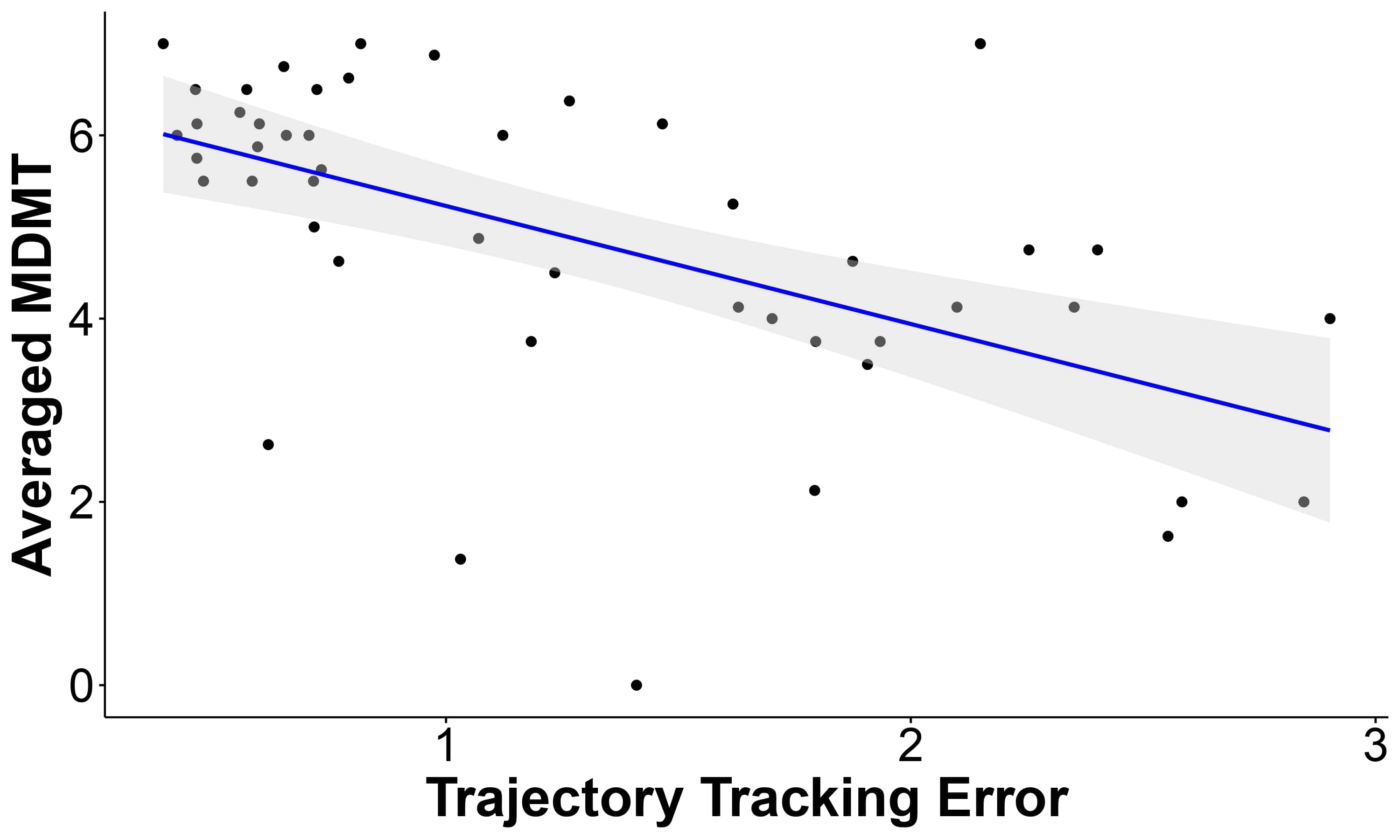}
        \caption{Trust and Tracking Error Correlation}
        \label{fig:perf_trust_corr}
    \end{subfigure}
    \hspace*{\fill}
    \begin{subfigure}[t]{0.235\textwidth}
        \centering
        \includegraphics[width=\textwidth]{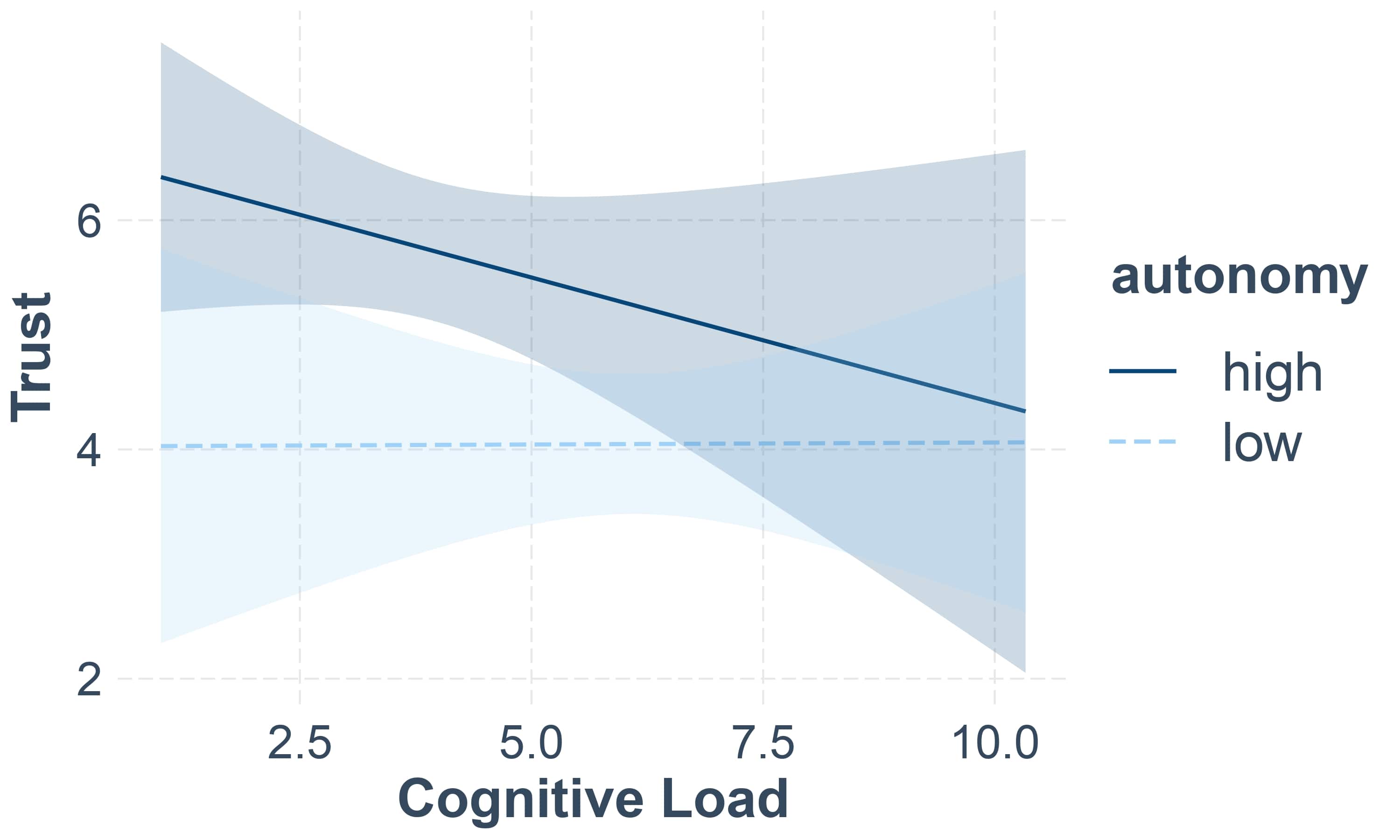}
        \caption{Three-Way Interaction Effect Plot}
        \label{fig:cl_trust_interact}
    \end{subfigure}
    \hspace*{\fill}
    
    \caption{\textbf{(a)-(f)} Plots of data distribution of each measure against \textit{autonomy}. Red bars show the \textit{mean} values of each distribution. \(\ast\) denotes $p<.05$. \textbf{(g)} Correlation between averaged \textit{MDMT} and \textit{trajectory tracking error}. \textbf{(h)} Illustration of interaction effect between \textit{cognitive load}, \textit{trust}, and robot \textit{autonomy}.}
    \label{fig:result_plots}
\end{figure*}

\subsection{Perceived Autonomy} \label{subsec:results_p_auto}
We evaluated whether the participants' perceived autonomy levels were consistent with the true autonomy levels. ANOVA indicated a clear \textit{autonomy level} effect on the \textit{perceived autonomy} (\(F(1, 22)=62.104, p < .001, \eta^2=.262\)). This suggests that participants correctly perceived the applied autonomy levels and serves as an experiment manipulation check.

\subsection{Cognitive Load} \label{subsec:results_cl}
We first investigated the effect of \textit{autonomy} on the three cognitive load measures: \textit{NASA-TLX} (Fig. \ref{fig:tlx}), \textit{tapping error} (Fig. \ref{fig:tapping_err_short}, \ref{fig:tapping_err_long}), and \textit{pupil diameter index} (Fig. \ref{fig:pupil_diameter}). The \textit{autonomy} level affected the raw \textit{NASA-TLX} score (\(F(1, 22)=46.615, p < .001, \eta^2=.262\)), where the 6 subscales showed internal consistency (Cronbach's \(\alpha = .762\)). In contrast to this result, no clear effect of \textit{autonomy} was observed on the \textit{tapping error} for both the short (\(F(1, 22)=1.137, p=.298, \eta^2=.028\)) and long (\(F(1, 22)=.873, p=.36, \eta^2=.008\)) rhythm intervals (tapping error mean and standard deviation is summarized in Table \ref{table:error}). Similarly, no clear \textit{autonomy} effect was observed on the \textit{pupil diameter index} (\(F(1, 22)=.088, p=.769, \eta^2=.0001\)). Therefore while the subjective NASA-TLX load assessment indicated that users perceived lower cognitive load when the robot had higher autonomy, the dual-task and physiological objective measures did not show the same effect.

\begin{table}[]
\centering
\begin{tabular}{lllll}
\textbf{}                  & \multicolumn{2}{l}{\textbf{Low Autonomy}} & \multicolumn{2}{l}{\textbf{High Autonomy}} \\ \toprule
\textbf{Measure}           & \textbf{$\mu$}          & \textbf{$\sigma$}         & \textbf{$\mu$}          & \textbf{$\sigma$}          \\ \midrule
Traj Tracking Error (cm)   & 2.163               & 0.402               & 0.834               & 0.069                \\
Tapping Error (short) (\%) & 7.458               & 9.615               & 5.513               & 6.768                \\
Tapping Error (long) (\%)  & 8.129               & 10.402              & 6.849               & 10.402               \\ \bottomrule
\end{tabular}
\caption{Mean ($\mu$) and standard deviation ($\sigma$) of the \textit{trajectory tracking} and \textit{tapping errors (short \& long)} under low and high \textit{autonomy} conditions.}
\label{table:error}
\end{table}

\subsection{Trust} \label{subsec:results_trust}
The \textit{MDMT} (Fig. \ref{fig:mdmt}) averaged across the \textit{Reliable} and \textit{Capable} sub-categories (Cronbach's \(\alpha=.968\)) indicated a clear autonomy effect (\(F(1, 22)=41.418, p<.001, \eta^2=.248\)). Similarly, the \textit{autonomy level} was also observed to affect the \textit{single-scale trust} measure (\(F(1, 22)=33.169, p<.001, \eta^2=.234\)). Here, Pearson correlation analysis between averaged \textit{MDMT} and data from the \textit{single-scale trust} question found these two measures to be strongly positively correlated (\(r(46) = .873\), \(p < .001\)). Overall, these results indicate that users had higher trust under the \textit{high autonomy} condition, despite having lower control authority in this condition.

\subsection{Teleoperation Task Performance} \label{subsec:results_tracking_err}
We first compared the participants' tracking performance in the \textit{baseline} and \textit{no autonomy} rounds. A paired t-test showed no clear difference between the two means (\(t(23) = -.429, p = .672, d = -.083\)), indicating that the secondary task did not impact the participants' primary task performance. The ANOVA showed an effect of \textit{autonomy} on \textit{trajectory tracking error} (Fig. \ref{fig:traj_err}, \(F(1, 22)=129.005, p<.001, \eta^2=.748\)), where participants performed trajectory tracking more accurately with a higher level of robot \textit{autonomy}.

We further investigated the effect of tracking performance on the level of trust in the robot. A Pearson correlation analysis between the averaged \textit{MDMT} measure and \textit{trajectory tracking error} found the two measures to be negatively correlated (\(r(46) = -.563\), \(p < .001\)) as shown in Figure \ref{fig:perf_trust_corr}. This suggests that participants adopted lower levels of trust towards the robot when they struggled to accurately track the trajectory.

\subsection{Cognitive Load \& Trust} \label{subsec:cl_and_trust}

We also explored the relationship between \textit{autonomy}, \textit{cognitive load} and \textit{trust}. A multiple regression showed a collective effect between \textit{cognitive load} (\textit{NASA-TLX}), \textit{autonomy} and \textit{trust} (average \textit{MDMT}), \((F(3, 44) = 5.734, p = .002, R^2 = 0.2811, R^2_{adj} = 0.232)\). However, there was no clear interaction between \textit{cognitive load} and \textit{autonomy} on \textit{trust} \((t = -.96, p = .342)\). Figure \ref{fig:cl_trust_interact} shows the relationship between \textit{cognitive load} and \textit{trust}, grouped by \textit{autonomy}. With high robot \textit{autonomy}, there was a potential trend of participants reporting lower \textit{trust} with higher \textit{cognitive load} and vice versa. Under low \textit{autonomy}, \textit{trust} levels seemed approximately constant across the spectrum of \textit{cognitive load} levels.

\section{Discussion} \label{sec:discussion}

\subsection{Impact of Autonomy on Subjective Cognitive Load Measure} \label{subsec:discussion1}
Our results show that the subjective \textit{cognitive load} is lower with higher robot \textit{autonomy}. This is consistent with \textbf{H1} (and previous results in other applications \cite{lin2020shared, young2019formalized}) and suggests that users perceived the robot's intervention as reducing their own workload. However, in contrast to the effects observed on the subjective measure, neither of the objective measures indicated a clear effect of \textit{autonomy} on \textit{cognitive load}. 

While all three methods have been shown to be effective at quantifying cognitive load, they each record at different time-scales - self-assessment is averaged over the entire round, tapping is recorded at the pattern frequency and pupil-diameter is recorded at 40Hz. Ideally for applications like teleoperation, cognitive load would be measured in real-time, enabling the robot's actions to be adapted in response to sudden load changes. Due to the discrepancy across measures, it is uncertain if our metrics are capturing different aspects of workload and of which is most appropriate for future real-time measurement. Here, a unifying framework across measures taken at multiple time-scales \cite{asselborn2019bridging} might enable clearer evaluation.

All of the tested methods also suffer either from individual bias or other confounding factors which are difficult to control for in a realistic teleoperation setting \cite{esmaeili2021current, strobach2015better}. For instance, participants repeatedly reported that high \textit{autonomy} made accurate trajectory tracking ``more manageable and interesting". This suggests that the design of dual-task setups may need to consider the user's cognitive resource allocation as a function of not only the task's cognitive demand, but also its engagement. Therefore, given the observed inconsistency in our results, determining which measure provides the ``ground-truth" of cognitive load remains an open problem.

\subsection{Autonomy Affects Trust Levels} \label{subsec:discussion2}
The \textit{MDMT} and the \textit{single-scale trust} questionnaires support \textbf{H2}, where participants' trust towards the robot increased with higher \textit{autonomy}. As the \textit{MDMT} assesses both the reliability and capability aspects of trust, this finding suggests that participants could recognize and understand the robot's behavior and identify its superior performance in the trajectory tracking task. These results are however likely to be feedback-reliant \cite{desai2013impact}, where the visual feedback in our teleoperation setup, which showed both the path and time information of the reference trajectory, could have enabled better understanding of the robot's actions. Such communication of the robot's behavior is likely critical in establishing trust in autonomy, and therefore should be further studied. Furthermore, while we have considered participants' individual expertise through the demographics items and evaluated potential learning effects through preliminary analysis, we believe that rising expertise in operating the robot and increasing task familiarity may still affect the user's trust and cognitive load, and therefore need to be considered in the design of effective shared control systems.

The strong correlation between the averaged \textit{MDMT} and the \textit{single-scale trust} measure suggests that a \textit{single-scale} measure may capture a subjective evaluation of trust as well as the \textit{MDMT} for this class of tasks. Here, the use of a \textit{single-scale} trust measure may also assist to develop higher frequency trust measures, as users can self-report more frequently. Indeed, using a similar single-scale trust question, \cite{ayoub2023real} collected binary trust information (trust / distrust) from participants in 25\,s time intervals in a simulated driving scenario, which was then used to adapt the assistance of an autonomous driving system. However, as trust is not a binary measure, this motivates further research into understanding its nuanced relationship with \textit{autonomy} to enable the design of finer-grained robot shared-control systems.

\subsection{Does Cognitive Load Affect Trust?} \label{subsec:discussion3}

Our results do not support \textbf{H3}, as we did not find an interaction between \textit{cognitive load} and \textit{trust} based on \textit{autonomy}. However, visual inspection of the three-way interaction plot (Figure \ref{fig:cl_trust_interact}) appears to suggest a potential trend in \textit{trust} across the \textit{cognitive load} values, and this \textit{trust - cognitive load} relationship is different under the high and low \textit{autonomy} conditions. The downwards trend in the \textit{high autonomy} condition indicates that when the robot has high control in the task, participants either have more \textit{trust} in the robot and experience less \textit{cognitive load}, or have less \textit{trust} and perceive the task to be more \textit{cognitively demanding}. In contrast with \textit{low autonomy}, where the human user exerted more effort into the task, their \textit{trust} in the robot remains approximately constant for all perceived levels of \textit{cognitive load}.

The absence of a three-way interaction effect may suggest that both cognitive load and trust are indeed measures which are independently influenced by the robot's autonomy. On the other hand, the potential correlation shown in Figure \ref{fig:cl_trust_interact} motivates further investigation using finer autonomy variations, which may produce alternative results that unveil the existence of more complex underlying relationships between robot autonomy and the human's cognitive load and trust.

\subsection{Generalization to Other Factors} \label{subsec:discussion4}
In this study, we designed the robot's autonomous controller to be highly accurate in its trajectory tracking. Therefore, as participants reported trust on the \textit{Performance} category of the \textit{MDMT}, it is possible that the observed effect of autonomy on trust might reflect participants being more comfortable with better robot performance, rather than them having more trust in the robot. It is also possible that the more accurate performance may lead to a feeling of less cognitive load since less consideration is given to improving accuracy, which resonates with similar findings in \cite{endsley1995out} which indicate that automated systems could lead to lower levels of effort exerted by the operator. Here, previous works have demonstrated that factors such as robot error \cite{desai2012error, salem2015faulty} and system transparency \cite{alonso2018system} can indeed impact the operator's state during the interaction. Therefore, future studies should consider imperfect robot behaviour. 

It is also worth noting that the shared control implementation in this study used the blending formulation typical to robotics literature. However, differing results may be observed for other methods of sharing control for which the allocation of autonomy may be different, such as through inferring human intent \cite{wang2021intent} or solving for the robot's control using optimization techniques \cite{selvaggio2021shared}. To better generalize our findings, other controller designs should also be investigated.

While the chosen task of trajectory tracking allowed us to obtain a fundamental understanding of robot autonomy's effect on human cognitive load and trust while minimizing the effects of confounding factors, it is possible that the results of this study may be task-specific, and different results may be obtained from other tasks and alternative setups. Furthermore, as the observed results are likely to be feedback-reliant, and specifically for the visual feedback used in this study, a limitation arises from participants finding it difficult to perceive the depth dimension of the RViz display. It is therefore worth investigating how the visual feedback used in this study or alternative visual information may impact the user's state and task performances.

Additionally, previous works have shown that providing haptic feedback to the operator in shared control settings can result in better task performance \cite{balachandran2020adaptive, abbink2012haptic,zhang2021haptic, thomas2023haptic}, especially in situations with added workload such as from a concurrent second task \cite{becker2022haptic}. While \cite{thomas2023haptic} also found that haptic feedback led to lower mental effort exerted by the user, from our results it is also possible that visual feedback alone may be sufficient to create the same effect. Therefore, future works should explore how the integration of visual and various types of haptic feedback under different levels of robot autonomy affect the task performance, the operator's state, and their perception of the autonomy level.

Finally while our setup was designed to replicate teleoperation, we minimized the delay between the operator's command and the robot's motion in our setup through high-frequency communication and tuned PID joint controllers. Here, delay may also have an effect on user perception and the task performance, especially in tasks such as trajectory tracking which require high precision under time constraints. Therefore, teleoperation setups with more realistic delay should be considered in future work.

\section{Conclusion}
This paper explored the relationship between robot autonomy and human cognitive load and trust during a collaborative HRI task. While our hypothesis of observing lower cognitive load under higher robot autonomy was confirmed via questionnaire data, the objective physiological and dual-task performance measures produced results that did not align with those from the subjective assessment. Additionally, our analysis revealed no clear interaction between trust and cognitive load, thus motivating future work to use a more continuous spectrum of autonomy levels instead of binary (high, low) classification to further investigate the relationship between robot autonomy and the user's cognitive load and trust. The effects of varying the robot's reliability or the control interface properties on the user's cognitive load and trust levels are also promising directions worth investigating.


\bibliographystyle{ieeetr}
\bibliography{references}


\end{document}